\documentclass[10pt,twocolumn,letterpaper]{article}

\usepackage[pagenumbers]{cvpr} 
\usepackage{overpic}
\usepackage{makecell}
\usepackage{adjustbox} 
\usepackage{multirow}
\usepackage{float}
\usepackage{bbding}

\usepackage[dvipsnames]{xcolor}

\definecolor{cvprblue}{rgb}{0.21,0.49,0.74}
\usepackage[pagebackref,breaklinks,colorlinks,citecolor=cvprblue]{hyperref}

\def\paperID{15222} 
\def\confName{CVPR}
\def\confYear{2024}

\title{FlashOcc: Fast and Memory-Efficient Occupancy Prediction via Channel-to-Height Plugin}

\author{
Zichen Yu$^{1}$ \quad Changyong Shu$^{2}$$\textsuperscript{\Envelope}$ \quad Jiajun Deng$^{3}$ \quad Kangjie Lu$^{2}$ \quad Zongdai Liu$^{2}$ \\
\quad Jiangyong Yu$^{2}$ \quad Dawei Yang$^{2}$ \quad Hui Li$^{2}$ \quad Yan Chen$^{2}$ \\
$^1$Dalian University of Technology, $^2$Houmo AI, $^3$University of Adelaide \\
{\tt\small yuzichen@mail.dlut.edu.cn,jiajun.deng@adelaide.edu.au.,} \\
{\tt\small \{changyong.shu,kangjie.lu,zongdai.liu,jiangyong.yu,dawei.yang,hui.li,yan.chen\}@houmo.ai}
\vspace{-15pt}
}

\begin{document}

\makeatletter
\vspace{-20pt}
\let\@oldmaketitle\@maketitle
\renewcommand{\@maketitle}{\@oldmaketitle
\centering
    \begin{overpic}[width=0.950\textwidth]{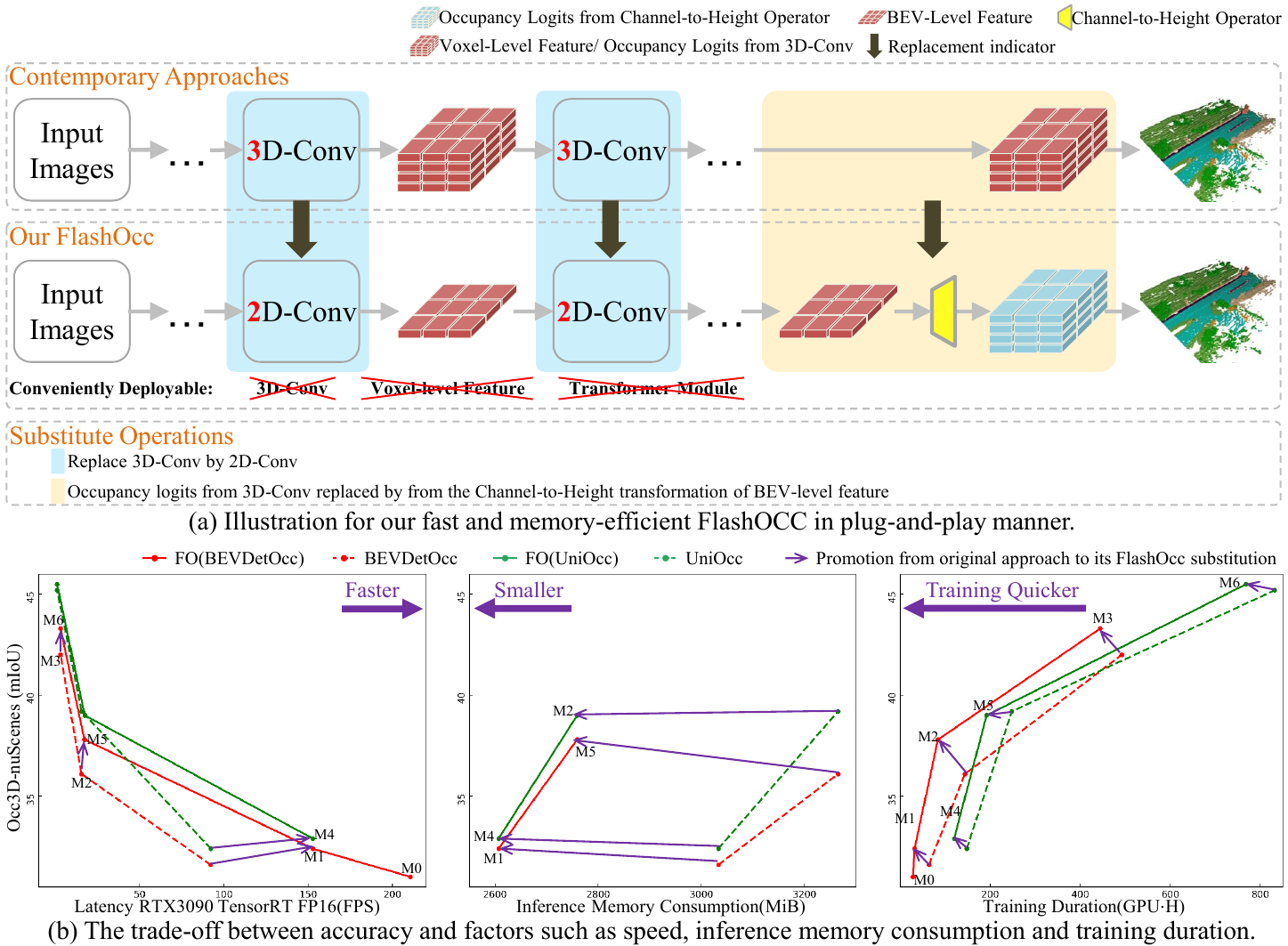}
    \end{overpic}
    \captionof{figure}{
    (a) illustrates how to achieve the proposed FlashOcc in plug-and-play manner. Contemporary approaches predict occupancy using voxel-level 3D feature processed by 3D-Conv.
    In contrast, our plugin substitute model achieves rapid and memory-efficient occupancy prediction by (1) replacing the 3D-Conv with 2D-Conv and (2) substituting the occupancy logits derived from the 3D-Conv with the Channel-to-Height transformation of BEV-level features acquired through 2D-Conv. 
    The abbreviation "Conv" represents convolution.
    (b) exemplifies the trade-off between accuracy and factors such as speed, inference memory consumption, and training duration. 
    The detailed configuration of M0-8 please consult to Table~\ref{tab:various_module_settings}.
    "FO" is an acronym that stands for FlashOcc, and "FO(*****)" represents the plugin substitution for the corresponding model named with "*****".  
    Best viewed in color.
    }
    \label{fig:fig1}
    \bigskip}                   
\makeatother

\maketitle
\begin{abstract}

Given the capability of mitigating the long-tail deficiencies and intricate-shaped absence prevalent in 3D object detection, occupancy prediction has become a pivotal component in autonomous driving systems.
However, the procession of three-dimensional voxel-level representations inevitably introduces large overhead in both memory and computation, 
obstructing the deployment of to-date occupancy prediction approaches.
In contrast to the trend of making the model larger and more complicated,
we argue that a desirable framework should be deployment-friendly to diverse chips while maintaining high precision.
To this end, we propose a plug-and-play paradigm, namely FlashOCC, to consolidate rapid and memory-efficient occupancy prediction while maintaining high precision. 
Particularly, our FlashOCC makes two improvements based on the contemporary voxel-level occupancy prediction approaches. Firstly, the features are kept in the BEV, enabling the employment of efficient 2D convolutional layers for feature extraction. Secondly, a channel-to-height transformation is introduced to lift the output logits from the BEV into the 3D space.
We apply the FlashOCC to diverse occupancy prediction baselines on the challenging Occ3D-nuScenes benchmarks and conduct extensive experiments to validate the effectiveness. The results substantiate the superiority of our plug-and-play paradigm over previous state-of-the-art methods in terms of precision, runtime efficiency, and memory costs, demonstrating its potential for deployment. The code will be made available.

\end{abstract}

\section{Introduction}
\label{sec:intro}
3D object detection in surround-view sense plays a crucial role in autonomous driving system.
Particularly, 
image-based 3D perception has received increasing attention from both academia and industry,
owing to its lower cost compared to LiDAR-dependent solutions and its promising performance~\cite{huang2021bevdet,li2022bevdepth,liu2022petr,shu20233DPPE,pointpillar,centerpoint}. 
Nevertheless, the 3D object detection task is limited to generate bounding boxes within predefined classes, which gives rise to two major challenges. Firstly, it encounters long-tail deficiencies, wherein unlabeled classes emerge in real-world scenarios beyond the existing predefined classes. Secondly, it faces the issue of intricate-shape absence, as complex and intricate geometry of diverse objects are not adequately captured by existing detection methods.

Recently, the emerged task of occupancy prediction addresses the aforementioned challenges by predicting the semantic class of each voxel in 3D space~\cite{li2023voxformer,huang2023tri,pan2023renderocc,pan2023uniocc}. This approach allows for the identification of objects that do not fit into the predefined categories and labels them as general objects. 
By operating at the voxel-level feature, these methods enable a more detailed representation of the scene, capturing intricate shapes and addressing the long-tail deficiencies in object detection.

The core of occupancy prediction lies in the effective construction of a 3D scene.
Conventional methods employ voxelization, where the 3D space is divided into voxels, and each voxel is assigned a vector to represent its occupancy status. Despite their accuracy, utilizing three-dimensional voxel-level representations introduces complex computations, including 3D (deformable) convolutions, transformer operators and so on\cite{zhang2023occformer,pan2023renderocc,sima2023_occnet,tian2023occ3d,occ-bev,Occupancy3ddet}. 
These pose significant challenges in terms of on-chip deployment and computational power requirements. 
To mitigate these challenges, sparse occupancy representation~\cite{wang2023panoocc} and tri-perspective view representation~\cite{huang2023tri} are investigated to conserve memory resources. However, this approach does not fundamentally address the challenges for deployment and computation.

Inspired by sub-pixel convolution techniques \cite{shi2016channel2spatial}, where image-upsampling is replaced by channel rearrangement, thus a Channel-to-Spatial feature transformation is achieved. 
Correspondly, in our work, we aim to implement a Channel-to-Height feature transformation efficiently. 
Given the advancement in BEV perception tasks, where each pixel in the BEV representation contains information about all objects in the corresponding pillar along height dimension, we intuitively utilize Channel-to-Height transformation for reshaping the flattened BEV features into three-dimensional voxel-level occupancy logits.  
Consequently, we focus on enhancing existing models in a general and plug-and-play manner, instead of developing novel model architectures, as listed in Figure.~\ref{fig:fig1} (a).
Detially, we direct replace the 3D convolution in contemporary methodologies with 2D convolution, and replacing the occupancy logits derived from the 3D convolution output with the Channel-to-Height transformation of BEV-level features obtained via 2D convolution.
These models not only achieves best trade-off between accuracy and time-consumption, but also demonstrates excellent deployment compatibility.

\section{Related Work}

\textbf{Voxel-level 3D Occupancy prediction.}
The earliest origins of 3D occupancy prediction can be traced back to Occupancy Grid Maps (OGM)~\cite{thrun2002probabilistic}, which aimed to extract detailed structural information of the 3D scene from images, and facilitating downstream planning and navigation tasks.
The existing studies can be classified into sparse perception and dense perception based on the type of supervision. The sparse perception category obtains direct supervision from lidar point clouds and are evaluated on lidar datasets~\cite{huang2023tri}.
Simultaneously, dense perception shares similarities with semantic scene completion (SSC)~\cite{armeni2017joint,dai2017scannet}.
Voxformer~\cite{li2023voxformer} utilizes 2.5D information to generate candidate queries and then obtains all voxel features via interpolation. Occ3D~\cite{tian2023occ3d} reformulate a coarse-to-fine voxel encoder to construct occupancy representation. RenderOcc~\cite{pan2023renderocc} extract 3D volume feature from surround views via 2D-to-3D network and predict density and label for each voxel with Nerf supervision. Furthermore, several benchmarks with dense occupancy labels are proposed~\cite{tian2023occ3d,sima2023_occnet}.
The approaches mentioned above voxelize 3D space with each voxel discribed by a vector~\cite{Occupancy3ddet,li2023fbocc,pan2023uniocc,sima2023_occnet,pan2023renderocc,cao2022monoscene,li2023voxformer}, as voxel-level representations with fine-grained 3D structure are inherently well-suited for 3D semantic occupancy prediction. 
However, the computational complexity and deployment challenges arised with voxel-based representations have prompted us to seek more efficient alternatives.

\textbf{BEV-based 3D Scene Perception.}
BEV-based methods employ a vector to represent the features of an entire pillar on BEV grid. 
Compared to voxel-based methods, 
it reduces feature representation in height-dimension for more computationally efficient, 
and also avoid the need for 3D convolutions for more deployment-friendly.
Promising results have demonstrated on diverse 3d scene perceptions, such as 3D lane detection~\cite{wang2023bevlanedet}, depth estimation\cite{wei2023surrounddepth}, 3D object detection~\cite{huang2021bevdet,li2022bevdepth,liu2022bevfusion} and 3D object tracking~\cite{zhou2022persdet}.
Although there are no methods performing occupancy prediction based on BEV-level features, however, BEV-level features can capture height information implicitly, which has been validated in scenarios of uneven road surfaces or suspended objects. These findings prompt us to leverage BEV-level features for efficient occupancy prediction.

\textbf{Efficient Sub-pixel Paradigm.}
The sub-pixel convolution layer first proposed in image super-resolution~\cite{shi2016channel2spatial} is capable of super-resolving low resolution data into high resolution space with very little additional computational cost compared to a deconvolution layer. 
The same idea has also been applied on BEV segmentation~\cite{liu2022petrv2}, 
wherein the segmentation representation of an $8\times8$ grid size is described by a segmentation query,
thus only 625 seg queries are used to predict the final $200\times200$ BEV segmentation results.
Based on the aforementioned approaches, 
we propose the Channel-to-Height transformation as an efficient method for occupancy prediction, wherein the occupancy logits are directly reshaped from the flattened BEV-level feature via the Channel-to-Height transform.
To the best of our knowledge, we are the pioneers in applying the sub-pixel paradigm to the occupancy task with utilizing BEV-level features exclusively, while completely eschewing the use of computational 3D convolutions.

\begin{figure*}
\centering
	\includegraphics[width=1.0\linewidth]{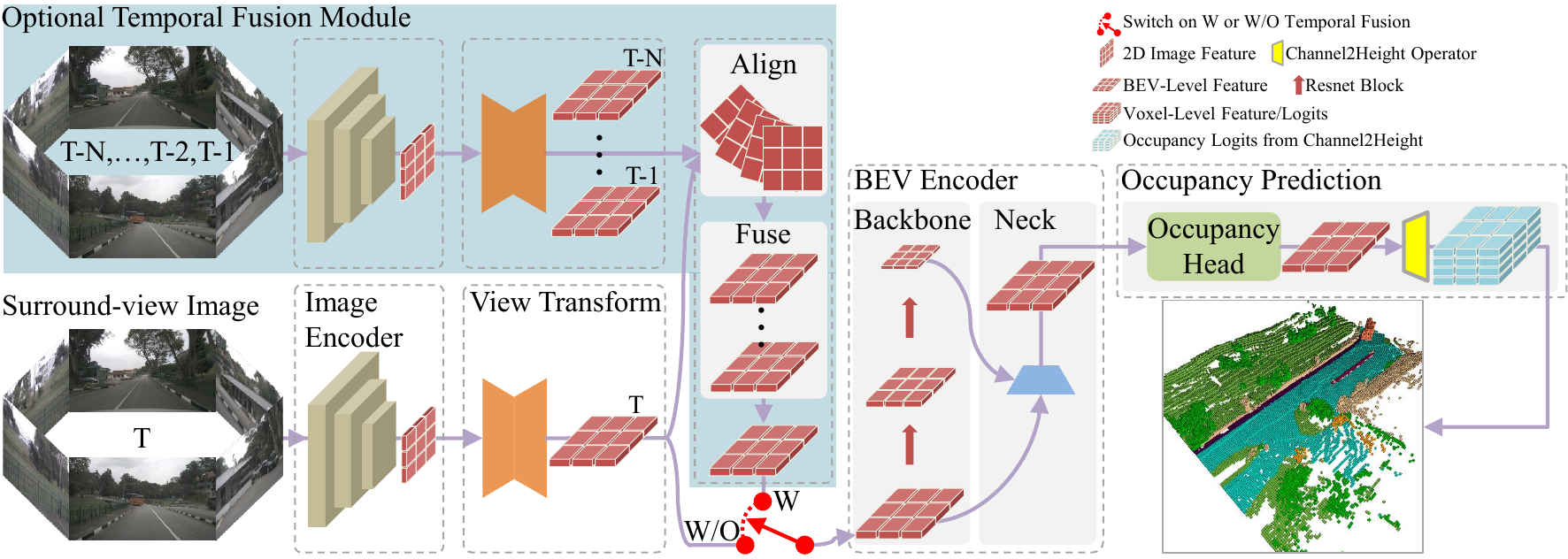}
	\caption{The diagram illustrates the overarching architecture of our FlashOcc, which is best viewed in color and with zoom functionality. The region designated by the dashed box indicates the presence of replaceable modules.
    The feature shapes of each replaceable module are denoted by icons representing 2D image, BEV-level, and voxel-level features, respectively.
    The light blue region corresponds to the optional temporal fusion module, and its utilization is contingent upon the activation of the red switch.
	}
	\label{fig:overview}
\end{figure*}

\section{Framework}
FlashOcc represents a pioneering contribution in the field by successfully accomplishing real-time surround-view 3D occupancy prediction with remarkable accuracy. 
Moreover, it exhibits enhanced versatility for deployment across diverse on-vehicle platforms, as it obviates the need for costly voxel-level feature procession, wherein view transformer or 3D (deformable) convolution operators are avoided. 
As denoted in Figure.~\ref{fig:overview}, the input data for FlashOcc consists of surround-view images, while the output is dense occupancy prediction results. 
Though our FlashOcc focuses on enhancing existing models in a general and plug-and-play manner, it can still be compartmentalized into five fundamental modules: (1) A 2D image encoder responsible for extracting image features from multi-camera images. (2) A view transformation module that facilitates the mapping from 2D perceptive-view image features into 3D BEV representation. (3) A BEV encoder tasked with processing the BEV feature information (4) Occupancy prediction module that predicts segmentation label for each voxel. (5) An optional temporal fusion module designed to integrate historical information for improved performance. 

\subsection{Image Encoder}\label{sec:image_view_encoder}
The image encoder extracts the input images to high-level features in perception-view.
Detailly, it utilizes a backbone network to extract multi-scale semantic features, which are subsequently fed into a neck module for fusion, thereby the semantic information with diverse granularities are fully exploited.
The classic ResNet~\cite{resnet} and strong SwinTransformer~\cite{liu2021Swin} is commonly chosen as the backbone network. 
ResNet's multiple residual-block design enables the elegant acquisition of feature representations with rich and multi-granularity semantic information.
Swin Transformer introduces a hierarchical structure that divides the input image into small patches and processes them in a progressive manner.
By utilizing a shifted window mechanism, SwinTransformer achieves high efficiency and scalability while maintaining competitive performance on various benchmarks.
As for the neck module, the concise FPN-LSS~\cite{huang2021bevdet,philion2020lift} was selected. It integrates the fine-grained features with directly upsampled coarse-grained features.
In fact, as the proposed paradigm that is never limited to a specific architecture, thus the backbone network can be replaced with other advanced models, such as SwinTransformer~\cite{liu2021Swin}, Vit~\cite{dosovitskiy2020vit}. And the neck module can also be substituted with other competitive variants, such as NAS-FPN~\cite{ghiasi2019fpn}, BiFPN~\cite{tan2020efficientdet}.

\subsection{View Transformer}\label{sec:view_transformer}
The view transformer is a crucial component in surround-view 3D perception system, it maps the 2D perceptive-view feature into BEV representation. Lift-splat-shot (LSS)~\cite{philion2020lift,huang2021bevdet} and Lidar Structure (LS)~\cite{li2023fast} have been widely used in recent work.
LSS leverages pixel-wise dense depth prediction and camera in/extrinsic parameters to project image features onto a predefined 3D grid voxels. Subsequently, pooling operations are applied along the vertical dimension (height) to obtain a flatten BEV representation. 
However, LS relies on the assumption of uniformly distributed depth to transfer features, which results in feature misalignment and subsequently causes false detections along camera-ray direction, though the computational complexity decreases.

\subsection{BEV Encoder}\label{sec:bev_encoder}
The BEV encoder enhances the coarse BEV feature obtained through view transformation, resulting in a more detailed 3D representation. 
The architecture of the BEV encoder resembles that of image encoder, comprising a backbone and a neck. We adopt the setting outlined in section \ref{sec:image_view_encoder}.
The issue of center features missing~\cite{fan2023super} (for LSS) or aliasing artifacts (for LS) is improved via feature diffusion after several blocks in the backbone.
As illustrated in Figure.~\ref{fig:overview}, two multi-scale features are integrated to enhance the representation quality.

\subsection{Occupancy Prediction Module}\label{sec:task_head}
As depicted in Figure.~\ref{fig:overview}, the BEV feature obtained from the neck for occupancy is fed into an occupancy head. 
It consists of a multi-layer convolutional network~\cite{bevdetocc, pan2023uniocc, pan2023renderocc} or complex multi-scale feature fusion module~\cite{li2023fbocc},
the latter exhibits a superior global receptive field, enabling a more comprehensive perception of the entire scene, while also providing finer characterization of local detailed features.
The resulting BEV feature from the occupancy head is then passed through the Channel-to-Height module. This module performs a simple reshape operation along the channel dimension, transforming the BEV feature from a shape of $B \times C \times W \times H$ to occupancy logits with a shape of $B \times C^{*} \times Z \times W \times H$, where $B$, $C$, $C^{*}$, $W$, $H$, and $Z$ represent the batch size, the channel number, the class number, the number of $x$/$y$/$z$ dimensions in the 3D space respectively, and $C=C^{*}\times Z$.

\subsection{Temporal Fusion Module}\label{sec:temporal_fusion_module}
The temporal fusion module is designed to enhance the perception of dynamic objects or attributes by integrating historical information. It consists of two main components: the spatio-temporal alignment module and the feature fusion module, as depicted in Figure~\ref{fig:overview}.
The alignment module utilizes ego information to align the historical BEV features with the current LiDAR system. This alignment process ensures that the historical features are properly interpolated and synchronized with the current perception system.
Once the alignment is performed, the aligned BEV features are passed to the feature fusion module. This module integrates the aligned features, taking into consideration their temporal context, to generate a comprehensive representation of the dynamic objects or attributes. The fusion process combines the relevant information from the historical features and the current perception inputs to improve the overall perception accuracy and reliability.


\section{Experiment}

\begin{table*}[h]
\footnotesize
\small
\setlength{\tabcolsep}{0.005\linewidth}

\def\mystrut{\rule{0pt}{1.5\normalbaselineskip}}
\centering
\begin{adjustbox}{width=2.1\columnwidth,center}
\begin{tabular}{l | c c | c | r r r r r r r r r r r r r r r r r r}
    \toprule[1.5pt]
    Method 
    & \rotatebox{90}{Backbone}
    & \rotatebox{90}{Image size}
    & \rotatebox{90}{\textbf{mIoU}$\uparrow$}  
    & \rotatebox{90}{\textbf{others}$\uparrow$} 
    & \rotatebox{90}{\textbf{barrier}$\uparrow$}
    & \rotatebox{90}{\textbf{bicycle}$\uparrow$} 
    & \rotatebox{90}{\textbf{bus}$\uparrow$} 
    & \rotatebox{90}{\textbf{car}$\uparrow$} 
    & \rotatebox{90}{\textbf{Cons. Veh}$\uparrow$} 
    & \rotatebox{90}{\textbf{motorcycle}$\uparrow$} 
    & \rotatebox{90}{\textbf{pedestrian}$\uparrow$} 
    & \rotatebox{90}{\textbf{traffic cone}$\uparrow$} 
    & \rotatebox{90}{\textbf{trailer}$\uparrow$} 
    & \rotatebox{90}{\textbf{truck}$\uparrow$} 
    & \rotatebox{90}{\textbf{Dri. Sur}$\uparrow$} 
    & \rotatebox{90}{\textbf{other flat}$\uparrow$} 
    & \rotatebox{90}{\textbf{sidewalk}$\uparrow$} 
    & \rotatebox{90}{\textbf{terrain}$\uparrow$} 
    & \rotatebox{90}{\textbf{manmade}$\uparrow$} 
    & \rotatebox{90}{\textbf{vegetation}$\uparrow$} 
    \\
    \midrule
    MonoScene~\cite{cao2022monoscene} & R101$\ast$ & 928$\times$1600 & $\textcolor{white}{0}$6.0 & 1.7 & 7.2 & 4.2 & 4.9 & 9.3 & 5.6 & 3.9 & 3.0 & 5.9 & 4.4 & 7.1 & 14.9 & 6.3 & 7.9 & 7.4 & 1.0 & 7.6 \\
    OccFormer~\cite{zhang2023occformer} & R101$\ast$ & 928$\times$1600 & 21.9 & 5.9 & 30.2 & 12.3 & 34.4 & 39.1 & 14.4 & 16.4 & 17.2 & 9.2 & 13.9 & 26.3 & 50.9 & 30.9 & 34.6 & 22.7 & 6.7 & 6.9 \\
    TPVFormer~\cite{huang2023tri} & R101$\ast$ & 928$\times$1600 & 27.8 & 7.2 & 38.9 & 13.6 & 40.7 & 45.9 & 17.2 & 19.9 & 18.8 & 14.3 & 26.6 & 34.1 & 55.6 & 35.4 & 37.5 & 30.7 & 19.4 & 16.7 \\
    CTF-Occ~\cite{tian2023occ3d} & R101$\ast$ & 928$\times$1600 & 28.5 & 8.0 & 39.3 & 20.5 & 38.2 & 42.2 & 16.9 & 24.5 & 22.7 & 21.0 & 22.9 & 31.1 & 53.3 & 33.8 & 37.9 & 33.2 & 20.7 & 18.0 \\
    RenderOcc~\cite{pan2023renderocc} & SwinB & 512$\times$1408 & 26.1 & 4.8 & 31.7 & 10.7 & 27.6 & 26.4 & 13.8 & 18.2 & 17.6 & 17.8 & 21.1 & 23.2 & 63.2 & 36.4 & 46.2 & 44.2 & 19.5 & 20.7 \\
    PanoOcc$\dagger$~\cite{wang2023panoocc} & R101$\ast$ & 432$\times$800$\textcolor{white}{0}$ & 36.6 & 8.6 & 43.7 & 21.6 & 42.5 & 49.9 & 21.3 & 25.3 & 22.9 & 20.1 & 29.7 & 37.1 & 80.9 & 40.3 & 49.6 & 52.8 & 39.8 & 35.8 \\
    PanoOcc$\dagger$~\cite{wang2023panoocc} & R101$\ast$ & 864$\times$1600 & 41.6 & 11.9 & 49.8 & 28.9 & 45.4 & 54.7 & 25.2 & 32.9 & 28.8 & 30.7 & 33.8 & 41.3 & 83.1 & 45.0 & 53.8 & 56.1 & 45.1 & 40.1 \\
    PanoOcc$\dagger$~\cite{wang2023panoocc} & R101$\bullet$ & 864$\times$1600 & 42.2 & 11.6 & 50.4 & 29.6 & 49.4 & 55.5 & 23.2 & 33.2 & 30.5 & 30.9 & 34.4 & 42.5 & 83.3 & 44.2 & 54.4 & 56.0 & 45.9 & 40.4 \\
    \midrule
    BEVDetOcc$\dagger$~\cite{bevdetocc} & SwinB & 512$\times$1408 & 42.0 & 12.1 & 49.6 & 25.1 & 52.0 & 54.4 & 27.8 & 27.9 & 28.9 & 27.2 & 36.4 & 42.2 & 82.3 & 43.2 & 54.6 & 57.9 & 48.6 & 43.5 \\
    UniOcc$\dagger$$\diamond$~\cite{pan2023uniocc} & SwinB & 640$\times$1600 & 45.2 & - & - & - & - & - & - & - & - & - & - & - & - & - & - & - & - & - \\
    \midrule
    \bf{FO(BEVDetOcc)$\dagger$:M3} & SwinB                    & 512$\times$1408 & 43.3 & 12.9 & 50.5 & 27.4 & 52.4 & 55.6 & 27.4 & 29.0 & 28.6 & 29.7 & 37.5 & 43.1 & 84.0 & 46.5 & 56.3 & 59.3 & 51.0 & 44.6 \\
    \bf{FO(UniOcc)$\dagger$$\diamond$:M6}    & SwinB & 640$\times$1600 & 45.5 & 14.3 & 52.4 & 33.9 & 52.5 & 56.5 & 32.3 & 33.3 & 34.6 & 35.3 & 39.6 & 44.1 & 84.6 & 48.7 & 57.9 & 61.2 & 49.7 & 42.7 \\

\bottomrule[1.5pt]
\end{tabular}
\end{adjustbox}
\vspace{-0.3cm}
\caption{3D occupancy prediction performance on the Occ3D-nuScenes valuation dataset. 
The symbol $\ast$ indicates that the model is initialized from the pre-trained FCOS3D backbone. 
"Cons. Veh" represents construction vehicle, and "Dri. Sur" is short for driveable surface.
"Train. Dur." is an abbreviation for training duration.
"Mem." represents memory consumption during inference.
$\bullet$ means the backbone is pretrained by the nuScense segmentation.
The frame per second (FPS) metric is evaluated using RTX3090, employing the TensorRT benchmark with FP16 precision.
"FO" is an acronym that stands for FlashOcc, and "FO(*****)" represents the plugin substitution for the corresponding model named with "*****".  
$\dagger$ denotes the performance is reported with utilization of camera mask during training.
The symbol $\diamond$ means the utilization of class-balance weight for occupancy classification loss.
} 
\label{table:sota_occ_eval}
\end{table*}

\begin{table*}[h]
\scriptsize
\setlength{\tabcolsep}{2.5mm}
\footnotesize
\setlength{\tabcolsep}{1.5mm}
\centering
\begin{tabular}{c|c|cc|c|cc|c|c}
\toprule[1.5pt]   
\multicolumn{1}{c|}{\multirow{2}{*}{N.}} & \multirow{2}{*}{Size} & \multicolumn{2}{c|}{Image Encoder} & \multirow{2}{*}{View Transform} & \multicolumn{2}{c|}{BEV Encoder} & \multirow{2}{*}{Occupancy Head} & \multirow{2}{*}{Temporal Module}              \\ \cline{3-4} \cline{6-7}
\multicolumn{1}{c|}{}                      &                             & \multicolumn{1}{c|}{Backbone} & Neck &                                 & \multicolumn{1}{c|}{Backbone} & Neck &   \\ 
\midrule
M0 & 256$\times$704$\textcolor{white}{0}$ & \multicolumn{1}{c|}{$\textcolor{white}{00}$R50} & FL-256 & LSS-64,200$\times$200,1.0 & \multicolumn{1}{c|}{3B-128-256-512} & FL-256 & MC-128-256-288 & - \\ 
M1 & 256$\times$704$\textcolor{white}{0}$ & \multicolumn{1}{c|}{$\textcolor{white}{00}$R50} & FL-256 & LSS-64,200$\times$200,0.5 & \multicolumn{1}{c|}{3B-128-256-512} & FL-256 & MC-256-512-288 & - \\ 
M2 & 256$\times$704$\textcolor{white}{0}$ & \multicolumn{1}{c|}{$\textcolor{white}{00}$R50} & FL-256 & LSS-64,200$\times$200,0.5 & \multicolumn{1}{c|}{3B-128-256-512} & FL-256 & MC-256-512-288 & Stereo4D \\ 
M3 & 512$\times$1408 & \multicolumn{1}{c|}{SwinB} & FL-256 & LSS-64,200$\times$200,0.5 & \multicolumn{1}{c|}{3B-128-256-512} & FL-256 & MC-256-512-288 & Stereo4D \\ 
\midrule
M4 & 256$\times$704$\textcolor{white}{0}$ & \multicolumn{1}{c|}{$\textcolor{white}{00}$R50} & FL-256 & LSS-64,200$\times$200,0.5 & \multicolumn{1}{c|}{3b-128-256-512} & FL-256 & MC-256-512-288 & - \\ 
M5 & 256$\times$704$\textcolor{white}{0}$ & \multicolumn{1}{c|}{$\textcolor{white}{00}$R50} & FL-256 & LSS-64,200$\times$200,0.5 & \multicolumn{1}{c|}{3b-128-256-512} & FL-256 & MC-256-512-288 & Stereo4D \\ 
M6 & 640$\times$1600 & \multicolumn{1}{c|}{SwinB} & FL-256 & LSS-64,200$\times$200,0.5 & \multicolumn{1}{c|}{3b-128-256-512} & FL-256 & MC-256-512-288 & Stereo4D \\ 
\midrule
M7 & 256$\times$704\textcolor{white}{0} & \multicolumn{1}{c|}{$\textcolor{white}{00}$R50} & FL-256 & \makecell[c]{F-VTM-64,200$\times$200,0.5 \\ B-VTM-1L-80-320} & \multicolumn{1}{c|}{3b-128-256-512} & FL-256 & MSO-(256,256,256)-128-256 & - \\ 
M8 & 512$\times$1408 & \multicolumn{1}{c|}{$\textcolor{white}{0}$R101} & FL-256 & \makecell[c]{F-VTM-64,200$\times$200,0.5 \\ B-VTM-1L-80-320} & \multicolumn{1}{c|}{3b-128-256-512} & FL-256 & MSO-(256,256,256)-128-256 & Mono-align-concat \\ 
\bottomrule[1.5pt]
\end{tabular}
\vspace{-0.3cm}
\caption{Detail settings for various methodologies. The suffix "-number" signifies the count of channels within this module, while "number$\times$number" denotes the size of image or feature. "3B" and "1L" are abbreviations for 3 bottleNeck and 1 transformer layer respectively.
"BE" is short for bevformer encoder.
"MC" represnets multi-convolution Head. "FL" is short for FPN LSS.
",number" indicates the resolution of depth bin.
F-VTM and B-VTM denotes forward projection and depth-aware backward projection in \cite{li2023fbocc} respectively.
MSO refers to the multi-scale occupancy prediction head described in \cite{li2023fbocc}, and the suffix "-(number,...,number)" indicates the list of channel number for the multi-scale input featuers. 
Stereo4D refers to the utilization of stereo volume to enhance the depth prediction for LSS, without incorporating BEV feature from previous frame. Mono-align-concat signifies the utilization of mono depth prediction for LSS, where the bev feature from the history frame is aligned and concatenated along the channel.
}
\label{tab:various_module_settings}
\end{table*}

In this section, we first detail the benchmark and metrics, as well as the training details for our FlashOcc in Section.~\ref{sec:exp_setup}. 
Then, Section.~\ref{sec:sota_exp} present the main results of our FlashOcc with fair comparison to other state-of-the-art methods on occupancy prediction.
After that, we conduct extensive ablative experiments to investigate the effectiveness of each component in our proposed FlashOcc in Section.~\ref{sec:exp_ablation}.

\subsection{Experimental Setup}\label{sec:exp_setup}
\textbf{Benchmark.}
We conducted occupancy on the Occ3D-nuScenes~\cite{tian2023occ3d} datasets.
The Occ3D-nuScenes dataset comprises 700 scenes for training and 150 scenes for validation. The dataset covers a spatial range of -40m to 40m along the X and Y axis, and -1m to 5.4m along the Z axis. The occupancy labels are defined using voxels with dimensions of $0.4m \times 0.4m \times 0.4m$ for 17 categories.
Each driving scene contains 20 seconds of annotated perceptual data captured at a frequency of 2 Hz. 
The data collection vehicle is equipped with one LiDAR, five radars, and six cameras, enabling a comprehensive surround view of the vehicle's environment.
As for evaluation metrics, the mean intersection-over-union (mIoU) over all classes is reported.

\begin{figure*}
\centering
	\includegraphics[width=1.0\linewidth]{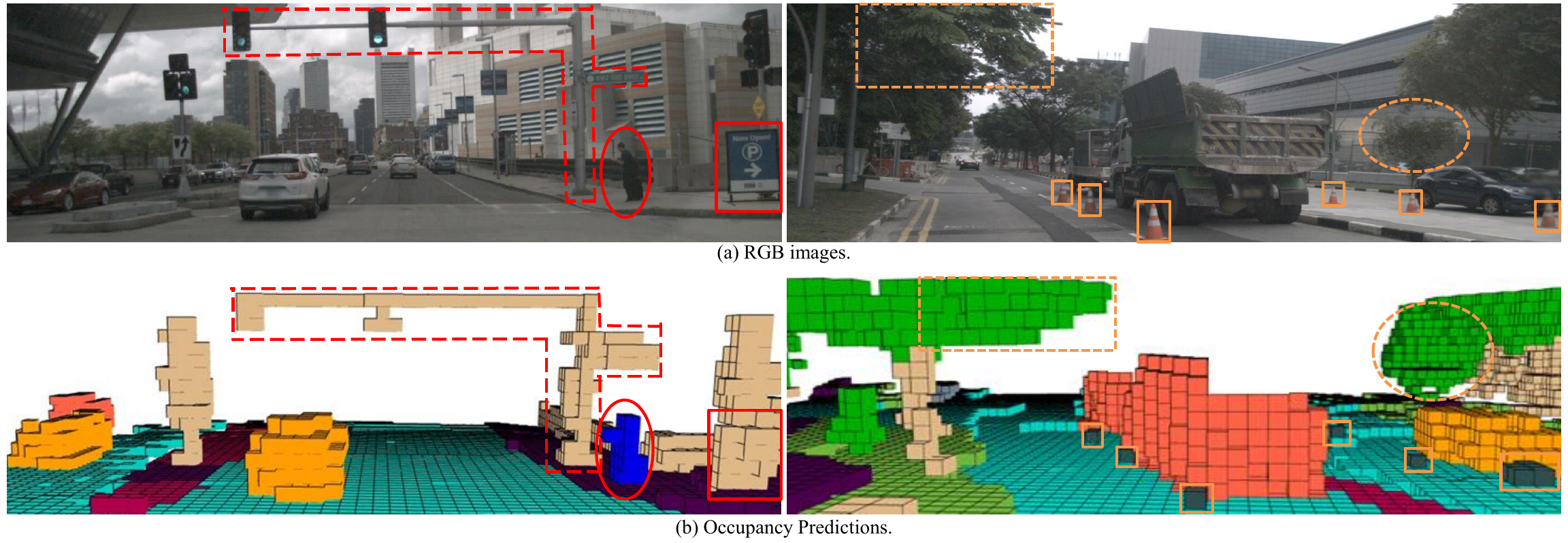}
    \vspace{-0.7cm}
	\caption{
 Qualitative results on Occ3D-nuScenes.
 Note that the perception range in Occ3D-nuScenes spans from -40m to 40m along the X and Y axes, and from -1m to 5.4m along the Z axis. Consequently, objects located outside of this range are not predicted.
	}
	\label{fig:visualization}
\end{figure*}

\textbf{Training Details.}
As our FlashOcc is designed in a plug-and-play manner, and the generalization and efficiency are demonstrated on diverse mainstream voxel-based occupancy methodologies, i.e. BEVDetOcc~\cite{bevdetocc}, UniOcc~\cite{pan2023uniocc} and FBOcc~\cite{li2023fbocc}. 
For a fair comparison, the training details are following the origin mainstream voxel-based occupancy methodologies strictly. 
As the channel number would be altered when replacing 3D convolution by 2D convolution, the detail architectures of respective plugin substitutions are presented in Table.~\ref{tab:various_module_settings}.
In the "Method" column of each experimental table, we use a ":" to associate each plugin substitution with its corresponding structure, i.e., M0-8.
All models are traind using the AdamW optimizer~\cite{loshchilov2017decoupled}, wherein a gradient clip is applied with learning rate 1e-4, with a total batch size of 64 distributed across 8 GPUS. The total training epoch for BEVDetOcc and UniOcc is set to 24, while FBOcc is trained for 20 epoch only.
Class-balanced grouping and sampling is not used in all experiments.

\subsection{Comparison with State-of-the-art Methods}\label{sec:sota_exp}

We evaluate our plugin FlashOcc on BEVDetOcc~\cite{bevdetocc} and UniOcc~\cite{pan2023uniocc},
and also compare the performance of our plugin substitutions with popular existing approaches, i.e. MonoScene~\cite{cao2022monoscene}, TPVFormer~\cite{huang2023tri}, OccFormer~\cite{zhang2023occformer}, CTF-Occ~\cite{tian2023occ3d}, RenderOcc~\cite{pan2023renderocc} and PanoOcc~\cite{wang2023panoocc}. 
As listed in Table.~\ref{table:sota_occ_eval}, 
3D occupancy prediction performances on the Occ3D-nuScenes valuation dataset are listed. Both the results with ResNet-101 and SwinTransformer-Base are evaluated. 
Our plug-and-play implementation of FlashOcc demonstrates improvement of 1.3 mIoU on BEVDetOcc. 
Additionally, the 0.3 mIoU enhancement on UniOcc further highlights the channel-to-height's ability to preserve voxel-level information within BEV feature, as the rendering supervision in UniOcc need fine-grained volume representation.
These results demonstrate the efficacy and generalizability of our proposed FlashOcc approach.
In addition, our FO(BEVDetOcc) surpasses the state-of-the-art transformer-based PanoOcc approach by 1.1 mIoU, further demonstrating the superior performance of our approach.

The qualitative visualization of FO(BEVDetOcc) is illustrated in Figure.~\ref{fig:visualization}, the traffic signal crossbar spanning over the road (indicated by the red dashed line) and the tree extending above the road (indicated by the orange dashed line) can both be effectively voxelized via our FO(BEVDetOcc), thus demonstrating the preservation of height information. With regards to the voxel description of pedestrians (indicated by the red ellipse), a forward protruding voxel at the chest signifies the mobile holded by the person, while the voxel extending behind the leg represents the suitcase pulled by the person. Furthermore, the small traffic cones are also observed in our predicted occupancy results (indicated by the solid orange rectangles). These findings collectively emphasize the outstanding capability of our FlashOcc in accurately capturing intricate shapes.

\subsection{Ablation Study}\label{sec:exp_ablation}
We conduct ablative experiments to demonstrate the efficacy of each component in our plugin substitution. 
Unless stated otherwise, 
all experiments employ ResNet-50 as the backbone network with a input image resolution of 704 x 256.
The spatial representation of 3D space is discretized into a grid size of 200 $\times$ 200 $\times$ 1.
The model are all pretrained on 3D object detection tasks.

\begin{figure}[H]
\centering
	\includegraphics[width=1.0\linewidth]{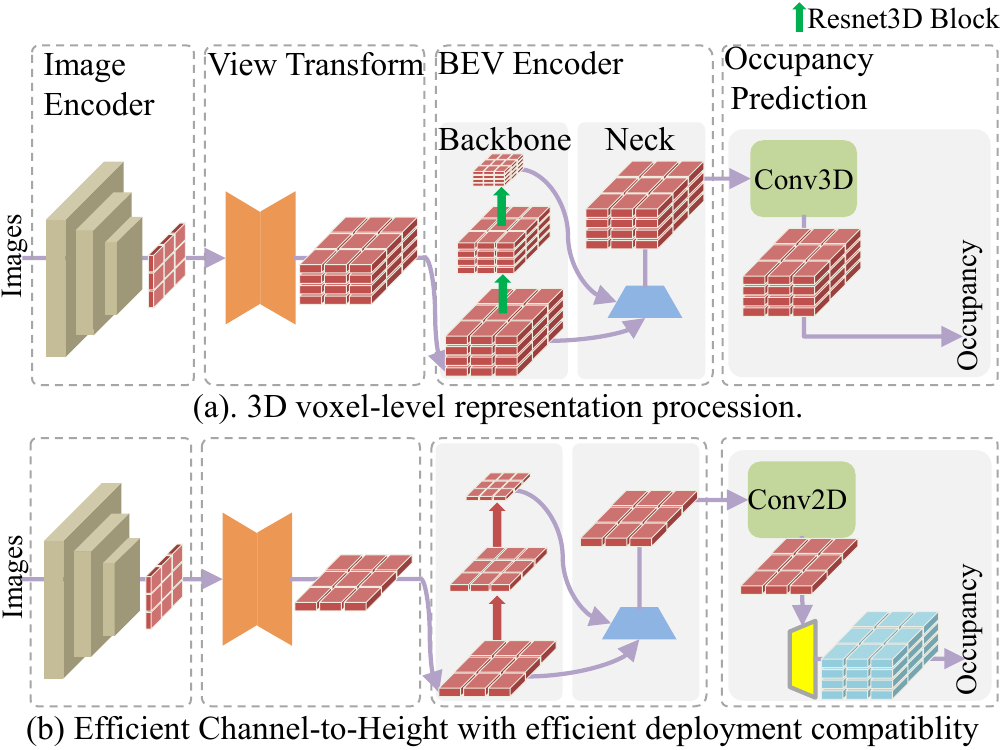}
    \vspace{-0.6cm}
	\caption{
    Architecture comparion between 3D voxel-level representation procession and ours plugin substitution.
    Apart from the instructions provided for the Resnet3D Block, all remaining icons comply with the guidelines presented in Figure~\ref{fig:fig1} and Figure~\ref{fig:overview}.
	}
    \label{fig:architecture_comparion_of_ours_and_ours_3d_extenstion}
\end{figure}

\textbf{Efficient Channel-to-Height Devoid of Complex 3D Convolution Computation.}\label{efficient_Channel-to-Height_devoid_of_3D_voxel_level_representation_procession}
We employ the Channel-to-Height operation at the output of the occupancy head, whereby the 2D feature is directly reshaped into 3D occupancy logits. This process does not involve explicit height-dimension representation learning.
From an intuitive standpoint, accurate 3D occupancy prediction necessitates a voxel-aware representation in three dimensions, involving complex 3D computations, as extensively discussed in prior research~\cite{sima2023_occnet,wang2023panoocc,pan2023renderocc}.
To ensure a fair comparison, we choose BEVDetOcc~\cite{bevdetocc} without temporal module as the voxel-level competitor.
As illustrated in Figure.~\ref{fig:architecture_comparion_of_ours_and_ours_3d_extenstion}. 
We decrease the grid size along the z-axis of the LSS to 1 and replace the 3D convolution in BEVDetOcc with 2D counterparts. Additionally, Channel-to-Height transformation is plugged at the output of the model.
The comparative results are presented in Table~\ref{fig:architecture_comparion_of_ours_and_ours_3d_extenstion}. 
Our M0 method, despite incurring a mere 0.6 mIoU performance degradation, achieves a speedup of more than twofold, surpassing the baseline method operated at 92.1 FPS with a rate of 210.6 FPS.
And our M1 module demonstrates superior performance, achieving a significant 0.8 mIoU improvement with a faster FPS of 60.6Hz compared to the 3D voxel-level representation approach. 
These outcomes further highlight the efficient deployment compatibility of our proposed Channel-to-Height paradigm, which eliminates the need for computational 3D voxel-level presentation procession. 

\begin{table}[htb] 
    \footnotesize
    \setlength{\tabcolsep}{6.0mm}
    \centering
    \begin{tabular}{l|c|c}
    \toprule[1.5pt]
     Method & mIoU & FPS \\ 
     \midrule
     3D Voxel-level Representation & 31.6 & $\textcolor{white}{0}$92.1 \\
     Ours:M0 & 31.0 & 210.6 \\
     Ours:M1 & 32.4 & 152.7 \\
    \bottomrule[1.5pt]
    \end{tabular}
    \vspace{-0.3cm}
    \caption{Comparison between 3D voxel-level representation procession and efficient Channel-to-Height. The FPS are test on RTX3090 by tensorrt with fp16 precision. 
    }
    \label{tab:Comparison_between_3D_voxel_level_representation_procession_and_efficient_Channel-to-Height}
\end{table}
\vspace{-0.8cm}

\vspace{-0.3cm}

\textbf{Generalizable FlashOcc on Diverse Methodologies.}\label{Generalizable_FlashOcc_on_Diverse_Methodologies}
In order to demonstrate the generalization of our plug-and-play FlashOcc, we aim to achieve convincing results by applying it on popular 3D convolution-based occupancy models like BEVDetOcc~\cite{bevdetocc}, RenderOcc~\cite{pan2023renderocc}, and FBOcc~\cite{li2023fbocc}. Specifically, we replace the 3D convolutions in these models with 2D convolutions, and substitute the occupancy logits obtained from the original model's final output with the occupancy logits obtained through the Channel-to-Height transformation.
The comparative results are presented in Table~\ref{tab:generalization_demonstration}, our method showcases superior performance.
Detailly, 
our plugin substitution, FO(BEVDetOcc), surpasses the original BEVDetOcc by 1.7 mIoU,
our FO(UniOcc) incurs a mere 0.2 mIoU performance degradation compared to the original UniOcc,
and our FO(FBOcc) acheves a 0.1 mIoU improvement compared to the origin FBOcc,
the aforementioned experimental results demonstrate significant improvements or remain comparable.
These findings across various methodologies provide further demonstration for the efficacy of our generalizable approach, which eliminates the requirement for computationally intensive 3D voxel-level presentation processing while ensures optimal performance.

\begin{table}[h] 
    \footnotesize
    \setlength{\tabcolsep}{12.5mm}
    \centering
    \begin{tabular}{l|c}
    \toprule[1.5pt]
     Method & mIoU \\ 
     \midrule
     BEVDetOcc\cite{bevdetocc}        & 36.1 \\
     FO(BEVDetOcc):M2                 & 37.8 \\
     \midrule
     UniOcc~\cite{pan2023uniocc}       & 39.2 \\
     FO(UniOcc):M5                     & 39.0 \\
     \midrule
     FBOcc~\cite{li2023fbocc}          & 37.2 \\
     FO(FBOcc):M8                      & 37.3 \\  
    \bottomrule[1.5pt]
    \end{tabular}
    \vspace{-0.3cm}
    \caption{
    Generalization demonstration of our plug-and-play FlashOcc on various popular voxel-level occupancy methodologies.
    The FPS are test on RTX3090 by tensorrt with fp16 precision. The abbreviation "FO" represent FlashOcc respectively. 
    }
    \label{tab:generalization_demonstration}
\end{table}
\vspace{-0.7cm}

\begin{table}[h] 
    \footnotesize
    \setlength{\tabcolsep}{10.0mm}
    \centering
    \begin{tabular}{l|c}
    \toprule[1.5pt]
     Method & mIoU \\ 
     \midrule
     BEVDetOcc(w/o T)~\cite{bevdetocc}  & 31.6{\fontsize{6}{14}\selectfont $\textcolor{white}{+0.0}$} \\
     BEVDetOcc~\cite{bevdetocc}         & 36.1{\fontsize{6}{14}\selectfont $\textcolor{red}{+4.5}$} \\
     FO(BEVDetOcc(w/o T)):M1            & 32.4{\fontsize{6}{14}\selectfont $\textcolor{white}{+0.0}$} \\
     FO(BEVDetOcc):M2                   & 37.8{\fontsize{6}{14}\selectfont $\textcolor{red}{+5.4}$} \\
     \midrule
     UniOcc(w/o T)~\cite{pan2023uniocc} & 32.4{\fontsize{6}{14}\selectfont $\textcolor{white}{+0.0}$} \\
     UniOcc~\cite{pan2023uniocc}        & 39.2{\fontsize{6}{14}\selectfont $\textcolor{red}{+6.8}$} \\
     FO(UniOcc(w/o T)):M4               & 32.9{\fontsize{6}{14}\selectfont $\textcolor{white}{+0.0}$} \\
     FO(UniOcc):M5                      & 39.0{\fontsize{6}{14}\selectfont $\textcolor{red}{+6.1}$} \\
     \midrule
     FBOcc(w/o T)~\cite{li2023fbocc}    & 32.7{\fontsize{6}{14}\selectfont $\textcolor{white}{+0.0}$} \\
     FBOcc~\cite{li2023fbocc}           & 37.2{\fontsize{6}{14}\selectfont $\textcolor{red}{+4.5}$} \\
     FO(FBOcc(w/o T)):M7                & 34.7{\fontsize{6}{14}\selectfont $\textcolor{white}{+0.0}$} \\
     FO(FBOcc):M8                       & 37.3{\fontsize{6}{14}\selectfont $\textcolor{red}{+2.6}$} \\
    \bottomrule[1.5pt]
    \end{tabular}
    \vspace{-0.3cm}
    \caption{
    Demonstration for consistent improvement in Temporal Module.
    "w/o T" denotes for without temporal module.
    }
    \label{tab:benfit_from_temporal_fusion}
\end{table}

\begin{table*}[h] 
    \footnotesize
    \setlength{\tabcolsep}{3.2mm}
    \centering
    \begin{tabular}{l|c|c|c|c|c|c|c}
    \toprule[1.5pt]
    \multirow{2}{*}{Method} & \multirow{2}{*}{FPS(Hz)} & \multicolumn{2}{c|}{Inference Duration(ms)}                               & \multicolumn{2}{c|}{Inference Memory(MiB)} & \multirow{1}{*}{Train. Dur.} & \multirow{1}{*}{Voxel-level}   \\ \cline{3-6}
     &  & \multicolumn{1}{c|}{Others} & \multicolumn{1}{c|}{BEV Enc.+Occ.} & \multicolumn{1}{c|}{Others} & \multicolumn{1}{c|}{BEV Enc.+Occ.} & {(GPU$\cdot$H)} & {Feature}\\ 
    \midrule
    BEVDetOcc(w/o T)~\cite{bevdetocc} & \textcolor{white}{0}92.1 & \textcolor{white}{0}3.4 & 7.5 & 2635 & 398 & \textcolor{white}{0}64 & \checkmark \\
    BEVDetOcc~\cite{bevdetocc}        & \textcolor{white}{0}15.5 & 57.0 & 7.5 & 2867 & 398 & 144 & \checkmark  \\
    FO(BEVDetOcc(w/o T)):M1           & 152.7 & \textcolor{white}{0}3.4 & 3.1 & 2483 & 124 & \textcolor{white}{0}32 & \scalebox{0.6}[1]{$\times$} \\
    FO(BEVDetOcc):M1                  & \textcolor{white}{0}17.3 & 54.7 & 3.1 & 2635 & 124 & \textcolor{white}{0}84 & \scalebox{0.6}[1]{$\times$} \\
     \midrule
    UniOcc(w/o T)~\cite{pan2023uniocc}   & \textcolor{white}{0}92.2 & \textcolor{white}{0}3.4 & 7.5 & 2635 & 398 & 148 & \checkmark \\
    UniOcc~\cite{pan2023uniocc}          & \textcolor{white}{0}15.6 & 57.0 & 7.5 & 2867 & 398 & 248 & \checkmark  \\
    FO(UniOcc(w/o T)):M4             & 152.7 & \textcolor{white}{0}3.4 & 3.1 & 2483 & 124 & 120 & \scalebox{0.6}[1]{$\times$} \\
    FO(UniOcc):M4                    & \textcolor{white}{0}17.5 & 54.7 & 3.1 & 2635 & 124 & 192 & \scalebox{0.6}[1]{$\times$} \\
    \bottomrule[1.5pt]
    \end{tabular}
    \vspace{-0.3cm}
    \caption{
    Analysis of Resource Consumption during training and deployment. The FPS are test on single RTX3090 by tensorrt with fp16 precision. 
    "Train. Dur." is short for training duration.
    "Enc.", "Occ." and "Feat" represent encoder, occupancy prediction and feature respectively. 
    "GPU$\cdot$H" denotes "1 GPU $\times$ 1 Hour". 
    }
    \vspace{-0.5cm}
    \label{tab:analysis_of_resource_consumption}
\end{table*}

\textbf{Consistent Improvement on Temporal Fusion.}\label{benefit_from_temporal}
Temporal augmentation is an essential tool in 3D perception for enhancing performance. To demonstrate the comparable performance of our plug-and-play FlashOcc before and after incorporating the temporal module compared to the original voxel-based approach, we conducted experimental validation using well-established temporal configurations of mainstream models as listed in Table.~\ref{tab:benfit_from_temporal_fusion}.
Compared to the baseline method BEVDetOcc, our FlashOcc exhibits improvements of 0.8 mIoU and 1.7 mIoU on both the non-temporal and temporal variants, respectively. Additionally, while the baseline method only achieves a 4.5 mIoU improvement when incorporating temporal information, our FlashOcc achieves a superior increase of 5.4 mIoU.
In term of the baseline method UniOcc, our FlashOcc achieves a 0.5 mIoU improvement on the non-temporal approach. And when temporal information is introduced, we observe a significant increase of 6.1 mIoU, this improvement aligns with the temporal enhancement observed in the baseline method.
As for the baseline method FBOcc, our FlashOcc achieves improvements of 2.0 mIoU and 0.1 mIoU on the non-temporal and temporal approaches, respectively. Moreover, in the temporal method, we observe an overall increase of 2.6 mIoU. However, the temporal improvement in our FlashOcc is not as significant as that of the baseline method. This is primarily due to the substantial improvement achieved by our non-temporal approach compared to the baseline method.
In conclusion, our FlashOcc demonstrates significant improvements when temporal information is introduced, compared to the non-temporal approach. Additionally, our FlashOcc achieves notable improvements or comparable performance compared to the baseline method in both the without and with temporal module configuration.

\textbf{Analyzation for Resource Consumption.}\label{analyzation_for_resource_consumption}
The performance of FlashOcc across diverse configurations has been validated in aforementioned paragraph, the resource consumption during model training and deployment will be further analyzed. Following the setting in Table.~\ref{tab:benfit_from_temporal_fusion}, we provide details on FPS, inference duration, inference memory consumption and training duration for each method.
Given the constrained applicability of our plugin, which exclusively impacts BEV encoder and occupancy head, we classify these two constituents as a distinct module to be examined. Meanwhile, the residual components, namely the image encoder and view transform, constitute a self-contained module referred to as "others" for analytical purposes.

In the case of BEVDetOcc, the utilization of our FlashOcc results in a notable reduction of 58.7\% in the inference duration of BEV encoder and occupancy prediction head, decreasing from 7.5 ms to 3.1 ms. At the same time, the inference memory consumption experiences a substantial savings of 68.8\% from 398 MiB to 124 MiB.
The training duration is reduced from 64 to 32 and from 144 to 84 respectively, for the experimental settings without and with temporal fusion module.
Moreover, owing to the temporal methodology implemented in BEVDetOcc is stereo matching, the "others" module exhibits notably longer inference time when operating in the temporal configuration. Nonetheless, the adoption of a channel-wise grouped matching mechanism results in a comparatively reduced memory overhead.
Similar conclusions were obtained on UniOcc as well, as it shares a similar model structure with BEVDetOcc.
However,  the integration of Rendering Supervision in UniOcc introduces a significant increase in training duration.

\section{Conclusion}
In this paper, we introduce a plug-and-play approach called FlashOCC, which aims to achieve fast and memory-efficient occupancy prediction. 
It directly replaces 3D convolutions in voxel-based occupancy approaches with 2D convolutions, and incorporates the Channel-to-Height transformation to reshape the flattened BEV feature into occupancy logits. 
The effectiveness and generalization of FlashOCC have been demonstrated across diverse voxel-level occupancy prediction methods. 
Extensive experiments have demonstrated the superiority of this approach over previous state-of-the-art methods in terms of precision, time consumption, memory efficiency, and deployment-friendly.
To the best of our knowledge, we are the first in applying the sub-pixel paradigm (Channel-to-Height) to the occupancy task with utilizing BEV-level features exclusively, completely avoiding the use of computational 3D (deformable) convolutions or transformer modules.
And the visualization results convincingly demonstrate that FlashOcc successfully preserves height information.
In our future work, we will explore the integration of our FlashOcc into the perception pipeline of autonomous driving, aiming to achieve efficient on-chip deployment.

\newpage
{
    \small
    \bibliographystyle{ieeenat_fullname}
    \bibliography{main}
}


\end{document}